\documentclass[runningheads]{llncs}
\usepackage{stmaryrd}

\usepackage{graphicx}
\usepackage[T1]{fontenc}
\usepackage[utf8]{inputenc}
\usepackage{comment}
\usepackage{amsmath,amssymb,mathtools}
\usepackage{float}
\usepackage{enumitem}                      
\usepackage{hyperref}   
\usepackage{algorithm}
\usepackage{algpseudocode}

\newcommand{\indep}{\mathrel{\perp\!\!\!\perp}}

\setlist[enumerate]{leftmargin=*,itemsep=2pt,topsep=2pt}

\usepackage{tikz}
\usetikzlibrary{positioning, arrows.meta, shapes.misc}
\begin{document}
\title{From Ethical Declarations to Provable Independence: An Ontology-Driven Optimal-Transport Framework for Certifiably Fair AI Systems}
%
%
\author{Sukriti Bhattacharya\inst{1} \and
Chitro Majumdar\inst{2}} 
%
\authorrunning{S. Bhattacharya et al.}
%
\institute{Senior Scientist, Trustworthy AI, Luxembourg Institute of Science \& Technology, Maison de l'innovation 5, L-4362, Luxembourg \\
\email{sukriti.bhattacharya@list.lu}
\and
Chief Investment Risk Strategist for Sovereign Institutions \\ Founder, RsRL, Jumeirah Beach Residence (JBR), P.O.Box 29215, Dubai, United Arab Emirates\\
\email{chitro.majumdar@rsquarerisklab.com}}
%
\maketitle              
\begin{abstract}

This paper introduces a novel framework for achieving provably fair AI systems that addresses the fundamental limitation of existing bias mitigation approaches: their inability to systematically identify and eliminate all forms of sensitive information, including subtle proxies that enable discriminatory outcomes through seemingly neutral features. Our integrated methodology leverages ontology engineering with OWL 2-QL to formally declare sensitive attributes and systematically discover their proxies through logical inference, compiling these into a comprehensive $\sigma$-algebra $G$ that represents the complete structure of biased measurable patterns. We then apply Delbaen–Majumdar optimal transport theory to construct new variables that are provably independent of $G$ while minimizing $L^2$-distance to preserve predictive accuracy. This approach transcends correlation-based debiasing techniques by providing exact independence guarantees rather than mere decorrelation. The framework's theoretical foundation rests on three key innovations: formalizing algorithmic bias as a structural property of learned $\sigma$-algebras, demonstrating systematic compilation of ontological knowledge into measure-theoretic structures, and establishing optimal transport as the unique solution for fair representation learning. Through detailed analysis of loan approval systems, we illustrate how ontology-guided $\sigma$-algebra construction captures indirect bias pathways (e.g., ZIP codes as proxies for race) that traditional methods overlook, ensuring comprehensive fairness while maintaining decision utility. The resulting methodology delivers a unified, certifiable approach to algorithmic fairness with applications spanning bias-free lending decisions and context-aware language model outputs, advancing the field from heuristic bias mitigation toward mathematically grounded fairness guarantees essential for trustworthy AI deployment in high-stakes societal applications.

\end{abstract}
\section*{Acknowledgment} \textit{The authors gratefully acknowledge Professor Emeritus \textbf{Dr. Freddy Delbaen} from the Department für Mathematik, ETH Zürich, Rämistrasse 101, 8092 Zürich, for his invaluable help, unwavering support, and insightful discussions and comments that significantly contributed to the development of this work.}
\section{Introduction}
Artificial intelligence (AI) systems, encompassing machine learning models and large language models (LLMs), are increasingly integral to high-stakes decision-making in domains such as financial lending, hiring, criminal justice, healthcare, and content generation. Yet, these systems frequently perpetuate algorithmic bias, producing unfair outcomes by learning and reproducing undesirable patterns embedded in historical training data. Such biases often stem from datasets that reflect societal inequities, where features like socio-economic status, race, gender, or their proxies (e.g., zip codes or employment history) inadvertently shape model decisions. For instance, automated loan approval systems trained on historical lending data may disproportionately deny applications from lower-income or minority groups, or offer them less favorable terms, despite comparable financial profiles. Similarly, as demonstrated by \cite{wang2025}, LLMs like Google Gemini have generated historically inaccurate depictions, such as racially diverse Nazis, while models like Claude have erroneously asserted uniform military fitness standards across genders. These errors highlight a critical flaw termed ``difference unawareness,'' where fair differentiation is conflated with harmful prejudice, leading to contextually inappropriate outcomes.

This issue is not merely technical but carries profound ethical and societal implications, demanding urgent and rigorous intervention. Algorithmic bias exacerbates existing disparities, undermines public trust in AI systems, and engenders severe ethical and legal consequences, including the perpetuation of systemic inequities and violations of human dignity. In practice, biased AI outcomes can deny individuals equitable access to critical opportunities—such as loans, jobs, or medical care—while entrenching modern forms of discrimination through ostensibly neutral ``color-blind'' or ``gender-blind'' approaches \cite{wang2025}. For example, enforcing uniform treatment in AI systems can backfire by ignoring contextual necessities, such as legal distinctions (e.g., gender-specific military drafts) or differential harm assessments (e.g., derogatory terms carrying greater weight against marginalized groups). These failures not only worsen societal outcomes but also erode the promise of AI as a force for progress. 

Addressing algorithmic bias is therefore paramount to ensuring fairness, promoting inclusivity, and aligning AI systems with ethical standards and regulatory frameworks. By developing principled methodologies that eliminate dependence on sensitive attributes while embracing \emph{contextual difference awareness}, we can create AI systems that deliver equitable, contextually sensitive, and trustworthy outcomes across diverse societal landscapes. Such an approach prevents AI from becoming a vector for division, instead harnessing its potential to advance societal good. To tackle this challenge, we propose a novel framework that integrates ontology engineering with measure-theoretic sigma-algebras to achieve provably fair AI systems. This approach conceptualizes bias as a structural property of the model's learned $\sigma$-algebra--the mathematical structure that encodes the ``measurable'' patterns and correlations derived from training data. By leveraging the declarative power of ontologies, we formally define sensitive attributes and their proxies (e.g., race, gender, or zip code) within a logical framework, specifically using OWL 2-QL for its computational efficiency and scalability in query answering \cite{horrocks2007description}. These ontological axioms are compiled into a sigma-algebra \( G \), which represents the undesirable information the model should not rely upon. Subsequently, we employ optimal transport techniques, building on advancements by \cite{delbaen2024approximation,delbaen2024convergence,RePEc:wsi:wschap:9789811280306_0009,Scandizzo_Majumdar}, to construct a new random variable \( Y \) that is independent of \( G \) and minimally distant from the original biased variable in the \( L^2 \)-norm. This ensures that the resulting representations are both fair (by achieving exact independence from sensitive attributes) and accurate (by preserving maximal information from the original data).

The remainder of this paper is structured as follows. Section \S\ref{sec:lr} reviews the state of the art on algorithmic bias mitigation, ontology-based frameworks, and measure-theoretic approaches, situating our contribution within ongoing debates on fairness. Section \S\ref{sec:sigma} introduces the $\sigma$-algebra formalism, establishing bias as a sub-$\sigma$-algebra and defining fairness through independence. Section \S\ref{sec:onto} develops the ontological foundation, presenting how sensitive attributes and their proxies are formally encoded. Section \S\ref{sec:onto_sigma} integrates these strands into the notion of ontology-guided $\sigma$-algebras, demonstrating their robustness and applying them to the loan approval case study. Section\S\ref{sec:imp}  outlines an implementation blueprint, mapping the theoretical framework into a reproducible pipeline for practice and auditability. Finally, Section \S\ref{sec:conclusion} concludes with key insights, emphasizing the theoretical and practical implications of our approach and identifying avenues for future research.

\section{Literature Review}
\label{sec:lr}

The growing deployment of artificial intelligence (AI) systems in high-stakes domains—such as financial lending, hiring, criminal justice, healthcare, and content generation—has amplified concerns about algorithmic bias, where models learn and perpetuate undesirable patterns from biased training data \cite{davis2025emerging}. These biases often stem from historical inequalities reflected in datasets, manifesting through sensitive attributes like race, gender, socio-economic status, or proxies (e.g., zip codes or employment history), leading to unfair outcomes that exacerbate societal disparities \cite{jain2024data,safranek2025ontological}. For instance, automated loan approval systems may disproportionately disadvantage marginalized groups, while clinical prediction models can exhibit fairness drift over time, widening performance gaps across subpopulations like race and sex \cite{delbaen2024convergence,delbaen2024approximation,davis2025emerging}.
Traditional bias mitigation strategies have primarily focused on correlation-based techniques, such as dataset balancing, fairness constraints, adversarial learning, and reweighting \cite{jain2024data}. However, these approaches often require group annotations, which are costly or unavailable, and may inadvertently reduce overall model accuracy or fail to address deeper structural biases \cite{jain2024data,zhou2025mitigating}. Recent critiques highlight the limitations of "difference-unaware" fairness, which enforces uniform treatment but overlooks contextual distinctions, potentially worsening inequities in applications requiring group-specific awareness \cite{wang2025}. Moreover, emerging issues like fairness drift—where post-deployment model fairness degrades due to evolving data distributions or clinical practices—underscore the need for ongoing monitoring and adaptive methods \cite{davis2025emerging}. In clinical settings, for example, random forest models for post-surgical outcomes have shown increasing gaps in accuracy and specificity between demographic groups over a decade, with model updates sometimes exacerbating rather than alleviating these disparities \cite{davis2025emerging}.

Ontology-based frameworks have gained traction for standardizing AI concepts and addressing ethical concerns, including bias. The Artificial Intelligence Ontology (AIO) provides a comprehensive hierarchy of AI methodologies, with a dedicated Bias branch encompassing 61 subclasses drawn from sources like NIST reports, categorizing biases across the AI lifecycle (e.g., computational, historical, societal) to promote fairness and transparency \cite{joachimiak2024artificial}. Similarly, knowledge graph-based ontological frameworks integrate predictive analytics, AI, and big data for decision-making systems, emphasizing ethical considerations such as algorithmic bias mitigation through diverse datasets, explainable AI (XAI), and compliance with regulations like the EU AI Act \cite{RePEc:wsi:wschap:9789811280306_0009}. These ontologies facilitate modular composition and ethical evaluations, aiding in bias identification and responsible AI deployment in sectors like healthcare and public policy \cite{joachimiak2024artificial,safranek2025ontological}.
Data-centric debiasing methods represent another evolving strand, focusing on identifying and removing problematic samples or shortcuts without extensive annotations. Data Debiasing with Datamodels (D3M) uses datamodeling to pinpoint training examples harming subgroup robustness, improving worst-group accuracy across datasets like CelebA and Waterbirds while preserving natural accuracy \cite{jain2024data}. Complementing this, Shortcut Hull Learning (SHL) formalizes shortcuts in a probabilistic measure-theoretic space, defining a "shortcut hull" as the minimal set of unintended features and employing diverse model suites to diagnose and eliminate them for unbiased evaluations \cite{zhou2025mitigating}. SHL's framework, validated on topological datasets, reveals insights into model capabilities (e.g., CNNs outperforming Transformers in global reasoning) and ensures independence from confounding correlations, enhancing fairness in high-dimensional AI assessments \cite{zhou2025mitigating}.
Building on these foundations, measure-theoretic approaches offer a principled path to achieving statistical independence from biased structures. Our prior works have pioneered this direction: \cite{delbaen2024approximation} and \cite{delbaen2024convergence} leverage optimal transport and Monge-Kantorovich problems to construct random variables independent of a sub-$\sigma$-algebra representing biased information, minimizing $L^2$-distance while ensuring stronger guarantees than mere decorrelation. Extending this, \cite{RePEc:wsi:wschap:9789811280306_0009} develops practical algorithms for discrete spaces, with applications to eliminating socio-economic dependencies in mortgage underwriting. In parallel, \cite{zhou2025mitigating} conceptualizes bias as a property of learned $\sigma$-algebras, proposing independence-based constructions to unify fair representation learning, counterfactual reasoning, and differential privacy.

Despite these advancements, significant gaps persist. Ontology frameworks, while useful for standardization, often lack integration with dynamic monitoring for fairness drift \cite{joachimiak2024artificial,davis2025emerging}. Data debiasing methods like D3M and SHL excel in annotation-scarce settings but may not scale to multidimensional cases or guarantee long-term independence from evolving biases \cite{jain2024data,zhou2025mitigating}. Measure-theoretic methods provide theoretical rigor but require further empirical validation in real-world AI pipelines \cite{delbaen2024approximation}. This review highlights the need for hybrid approaches that combine ontological structuring, data-centric debiasing, drift monitoring, and measure-theoretic independence to foster equitable, sustainable AI systems.

\section{The $\sigma$-Algebra: A Formalization of Observable Information}
\label{sec:sigma}

\subsubsection{Probability Space and Measurable Events}
A \emph{probability space} is a triple $(\Omega,\mathcal{F},\mathbb{P})$, where:
\begin{itemize}
\item The \textbf{sample space} $\Omega$ is a non-empty set of all possible outcomes. In our context, this can be a dataset of individual data points.
\item $\mathcal{F}$ is a \textbf{$\sigma$-algebra} on $\Omega$, which represents the collection of all events about which we can reason probabilistically.
\item $\mathbb{P}$ is a \textbf{probability measure} on $\mathcal{F}$, a function that assigns a likelihood (a value between 0 and 1) to each event in $\mathcal{F}$.
\end{itemize}

\subsection{The $\sigma$-Algebra: A Formalization of Observable Information}
A \emph{$\sigma$-algebra} $\mathcal{F}$ on $\Omega$ is a collection of subsets of $\Omega$ that represents the \emph{observable information}, i.e., all events that can be assigned a probability. Formally, $\mathcal{F} \subseteq 2^\Omega$ must satisfy:
\begin{enumerate}[label=(\roman*)]
    \item $\Omega \in \mathcal{F}$ (the sample space itself is measurable);
    \item If $A \in \mathcal{F}$, then its complement $A^{c} = \Omega \setminus A$ is also in $\mathcal{F}$;
    \item If $A_{1}, A_{2}, \ldots \in \mathcal{F}$, then $\bigcup_{k=1}^{\infty} A_{k} \in \mathcal{F}$.
\end{enumerate}
The pair $(\Omega,\mathcal{F})$ is called a \emph{measurable space}.

\subsection{Probability Measure}
A \emph{probability measure} is a mapping $\mathbb{P}: \mathcal{F} \to [0,1]$ such that:
\begin{enumerate}[label=(\alph*)]
    \item $\mathbb{P}(\Omega) = 1$;
    \item For any countable sequence of pairwise disjoint events $A_k \in \mathcal{F}$,
    \[
        \mathbb{P}\!\left( \bigcup_{k=1}^{\infty} A_k \right) = \sum_{k=1}^{\infty} \mathbb{P}(A_k).
    \]
\end{enumerate}
The triple $(\Omega,\mathcal{F},\mathbb{P})$ thus forms a \emph{probability space}.

\subsection{Random Variables and Bias as a Sub-$\sigma$-Algebra}

A \emph{random variable} is a measurable function $X: \Omega \to \mathbb{R}^{d}$ for some dimension $d \ge 1$. A function is measurable if for every Borel set $B \subseteq \mathbb{R}^{d}$ (the smallest $\sigma$-algebra generated by all open sets in $\mathbb{R}^d$), its preimage $X^{-1}(B)$ is an event in $\mathcal{F}$.
Within our framework, bias is conceptualized as a structural property of a model’s learned information. This information is encoded in a sub-$\sigma$-algebra $\mathcal{G} \subseteq \mathcal{F}$, which is the smallest $\sigma$-algebra containing all events related to sensitive attributes (e.g., race, gender, ZIP code) and their proxies. Intuitively, $\mathcal{G}$ captures all information a model could infer about protected characteristics.

\subsection{Independence from a $\sigma$-Algebra for Fairness}
A random variable $Y$ (e.g., a prediction or representation) is said to be \emph{independent of a $\sigma$-algebra $\mathcal{G}$} if and only if, for every Borel set $B \subseteq \mathbb{R}^d$ and every $G \in \mathcal{G}$,
\[
    \mathbb{P}(Y \in B \mid G) = \mathbb{P}(Y \in B).
\]
This definition is a direct mathematical embodiment of fairness: conditioning on any information in $\mathcal{G}$ does not alter the probability distribution of $Y$. This requirement is significantly stronger than zero correlation, as it mandates that the model reveals \emph{no information} about protected attributes.

\subsection{Best $L^{2}$-Approximation Independent of $\mathcal{G}$}
Given a potentially biased random variable $X$, the goal is to construct a fair representation $Y$ that is independent of $\mathcal{G}$ and optimally approximates $X$. The notion of optimality is formalized as minimizing the mean squared error in the Hilbert space $L^{2}(\Omega,\mathcal{F},\mathbb{P})$. Formally, we seek $Y \in L^{2}(\Omega,\mathcal{F},\mathbb{P})$ such that:
\begin{enumerate}[label=(\arabic*)]
    \item $Y$ is independent of $\mathcal{G}$;
    \item $Y$ minimizes
    \[
        \mathbb{E}\!\left[ \| X - Y \|^{2} \right] 
        = \inf_{Z \indep \mathcal{G}} \, \mathbb{E}\!\left[ \| X - Z \|^{2} \right].
    \]
\end{enumerate}
As established in \cite{delbaen2024approximation,delbaen2024convergence,RePEc:wsi:wschap:9789811280306_0009}, a unique solution $Y$ exists under the mild condition that $(\Omega,\mathcal{F},\mathbb{P})$ is \emph{atomless conditionally on $\mathcal{G}$}. The construction leverages optimal transport theory, solving for the optimal distribution and subsequently realizing it as a random variable independent of $\mathcal{G}$.

\subsection{A Case Study: Algorithmic Bias in Loan Approval}
To ground our theoretical framework in a socially consequential domain  \cite{kelly2021algorithmic,adegoke2024evaluating,ongena2016gender}, we consider the problem of \emph{loan approval}. Credit scoring and underwriting systems directly influence access to financial resources and opportunities, yet numerous studies and legal cases have revealed systematic disparities in approval rates across gender, racial, and socioeconomic lines. Such disparities often arise not because sensitive attributes are explicitly included in the model, but because correlated features act as \emph{proxies} for protected characteristics. This makes loan approval a paradigmatic example of algorithmic bias, and a suitable case study for the framework we develop in this work.

\subsubsection{Problem Formulation in a Probability Space}
We formalize the loan approval process as a probability space $(\Omega,\mathcal{F},\mathbb{P})$:
\begin{itemize}
    \item The \textbf{sample space} $\Omega$ is the set of all potential loan applicants. Each element $\omega \in \Omega$ corresponds to one applicant with a complete set of characteristics.
    \item The \textbf{$\sigma$-algebra} $\mathcal{F}$ is the collection of events measurable by the underwriting process. These events encode information about an applicant's credit history, income, debt-to-income ratio, employment record, and other relevant features.
    \item The \textbf{probability measure} $\mathbb{P}$ encodes the distribution of applicants in the population. In practice, $\mathbb{P}$ corresponds to the empirical distribution over a historical dataset of loan applicants.
\end{itemize}

\subsubsection{Random Variables and the $\sigma$-Algebra of Bias}
Within this probability space, the following random variables capture the structure of the problem:
\begin{itemize}
    \item $X: \Omega \to \mathbb{R}^d$ denotes the \textbf{original feature vector} available to the lender's model, consisting of variables such as income, credit score, debt-to-income ratio, and employment length.
    \item $S: \Omega \to \{ \text{male}, \text{female} \}$ represents a \textbf{sensitive attribute}, here gender. The information content of $S$ generates a sub-$\sigma$-algebra $\mathcal{G} \subseteq \mathcal{F}$ that includes not only gender itself, but also any measurable event that acts as a proxy (e.g., certain educational backgrounds, given names, or residential histories).
    \item $Y$ is the \textbf{fair representation} of an applicant’s financial profile, constructed to be independent of $\mathcal{G}$ while remaining the closest possible approximation to $X$ in the $L^{2}$ sense.
\end{itemize}

The central challenge is that even if the lender does not explicitly include $S$ in $X$, correlated variables may allow the model to reconstruct gender information, leading to discriminatory outcomes. By requiring $Y \indep \mathcal{G}$, our framework guarantees that no event in $\mathcal{G}$—including both $S$ and its proxies—can influence the distribution of $Y$. This independence criterion enforces a stringent mathematical notion of fairness, ensuring that decisions derived from $Y$ cannot be biased by sensitive information. 

In subsequent sections, we will leverage \emph{ontologies} to systematize the identification of proxies for sensitive attributes, thereby providing a principled method to specify the sub-$\sigma$-algebra $\mathcal{G}$ in complex domains. This case study will serve as a running example throughout the paper.

\section{Ontologies: A Formal Foundation for Knowledge Representation}
\label{sec:onto}

In our framework, we employ \emph{ontologies} to provide a formal and explicit specification of domain knowledge. An ontology is a logical theory that formally defines a shared conceptualization of a domain \cite{gruber1993translation,horrocks2007description}. Unlike a simple data schema, an ontology provides a rich set of logical axioms that enable automated reasoning and inference. We adopt a view of ontologies grounded in \emph{Description Logic (DL)}, a decidable fragment of first-order logic widely used in knowledge representation.

\subsection{Formal Components}
An ontology $\mathcal{O}$ is a tuple
\[
    \mathcal{O} = (\mathcal{C}, \mathcal{R}, \mathcal{I}, \mathcal{A}),
\]
where:
\begin{itemize}
    \item $\mathcal{C}$ is a finite set of \emph{class symbols} (concepts);
    \item $\mathcal{R}$ is a finite set of \emph{role symbols} (binary relations or properties);
    \item $\mathcal{I}$ is a finite set of \emph{individual symbols};
    \item $\mathcal{A}$ is a finite set of \emph{axioms}.
\end{itemize}

The meaning of these symbols is defined by an \emph{interpretation} 
\[
    \mathcal{M} = (\Delta^{\mathcal{M}}, \cdot^{\mathcal{M}}),
\]
where $\Delta^{\mathcal{M}}$ is a nonempty set called the \emph{domain of interpretation}, and $\cdot^{\mathcal{M}}$ is an interpretation function mapping symbols to elements and relations in this domain:
\begin{itemize}
    \item Each individual $a \in \mathcal{I}$ is mapped to an element $a^{\mathcal{M}} \in \Delta^{\mathcal{M}}$;
    \item Each class $C \in \mathcal{C}$ is mapped to a subset $C^{\mathcal{M}} \subseteq \Delta^{\mathcal{M}}$;
    \item Each role $r \in \mathcal{R}$ is mapped to a binary relation $r^{\mathcal{M}} \subseteq \Delta^{\mathcal{M}} \times \Delta^{\mathcal{M}}$.
\end{itemize}

In our setting, the domain $\Delta^{\mathcal{M}}$ corresponds directly to the sample space $\Omega$ from the probability space formulation. Thus, we can formally state the equivalence:
\[
    \Delta^{\mathcal{M}} \equiv \Omega.
\]

\subsection{Knowledge Base as an Axiomatic System}
An ontology’s knowledge is encapsulated in a set of logical axioms $\mathcal{A}$, typically partitioned into two components: the \emph{TBox} and the \emph{ABox}.
\begin{itemize}
    \item The \textbf{TBox} (Terminological Box), denoted $\mathcal{T}$, contains axioms that define relationships between classes and roles. These are general statements about the domain. For example, a subsumption axiom $C \sqsubseteq D$ is satisfied if $C^{\mathcal{M}} \subseteq D^{\mathcal{M}}$ for every interpretation $\mathcal{M}$, meaning that every instance of class $C$ is also an instance of class $D$.
    \item The \textbf{ABox} (Assertional Box), denoted $\mathcal{A}$, contains axioms about specific individuals. These are factual statements about the domain. For example, a class assertion $C(a)$ is satisfied if $a^{\mathcal{M}} \in C^{\mathcal{M}}$, and a role assertion $r(a,b)$ is satisfied if $(a^{\mathcal{M}}, b^{\mathcal{M}}) \in r^{\mathcal{M}}$.
\end{itemize}

Reasoning over an ontology involves determining whether a given axiom is a logical consequence of the knowledge base. This is denoted
\[
    \mathcal{O} \models \alpha,
\]
meaning that the axiom $\alpha$ is true in all interpretations that satisfy the axioms in $\mathcal{O}$.

\subsection{A Case Study: Ontological Modeling of Loan Approval Bias}
To illustrate how ontologies provide a rigorous foundation for capturing algorithmic bias, we revisit the loan approval problem using the formal components of an ontology. Recall that an ontology is defined as a tuple
\[
    O = (C, R, I, A),
\]
where \(C\) is a set of classes, \(R\) a set of roles, \(I\) a set of individuals, and \(A\) a set of axioms.

\begin{itemize}
    \item \textbf{Individuals} (\(I\)): Each loan applicant corresponds to an individual symbol. For example, \(i_1 = \texttt{JohnDoe}\) and \(i_2 = \texttt{JaneSmith}\).
    \item \textbf{Classes} (\(C\)): We define domain-relevant classes such as
    \[
        \texttt{LoanApplicant}, \quad \texttt{HighRiskApplicant}, \quad \texttt{ApprovedApplicant}, \quad \texttt{ProxyForLowIncome}.
    \]
    \item \textbf{Roles} (\(R\)): These encode binary relations, for example:
    \[
        \texttt{hasCreditScore}, \quad \texttt{livesInZIP}, \quad \texttt{hasEmploymentStatus}.
    \]
\end{itemize}

The semantics are given by an interpretation \(M = (\Delta^M, \cdot^M)\), where the domain \(\Delta^M\) corresponds to the set of all loan applicants (i.e., \(\Delta^M \equiv \Omega\) from the probability space formulation). The interpretation function \(\cdot^M\) maps:
\begin{itemize}
    \item an individual symbol \(a \in I\) to an element \(a^M \in \Delta^M\),
    \item a class \(C \in C\) to a subset \(C^M \subseteq \Delta^M\),
    \item a role \(r \in R\) to a binary relation \(r^M \subseteq \Delta^M \times \Delta^M\).
\end{itemize}

The knowledge base consists of axioms \(A = T \cup A'\), where \(T\) is the TBox and \(A'\) is the ABox:
\begin{itemize}
    \item \textbf{TBox axioms} define structural relationships. For instance:
    \[
        \exists \texttt{livesInZIP}.\{\texttt{ZIP\_12345}\} \sqsubseteq \texttt{ProxyForLowIncome}, 
        \texttt{ProxyForLowIncome} \sqsubseteq \texttt{SensitiveAttribute}.
    \]
    This encodes that applicants living in ZIP code 12345 belong to a proxy class for low income, which is classified as sensitive.
    \item \textbf{ABox axioms} describe assertions about individuals. For example:
    \[
        \texttt{LoanApplicant}(\texttt{JohnDoe}),
        \texttt{hasCreditScore}(\texttt{JohnDoe},580), 
        \texttt{livesInZIP}(\texttt{JohnDoe},\texttt{ZIP\_12345}).
    \]
\end{itemize}

Reasoning over this ontology reveals that:
\[
    \texttt{JohnDoe} \in (\texttt{ProxyForLowIncome})^M,
\]
which means that the applicant is inferred to be associated with a sensitive proxy. If this information is used in loan decision-making, the system can be flagged as exhibiting bias. 

Thus, the ontological framework provides a precise, formal mechanism for :
\begin{enumerate}
\item identifying proxies of sensitive attributes, \item enforcing fairness constraints via axioms, and \item ensuring that the reasoning process remains transparent and explainable.
\end{enumerate}

\section{Ontology-Guided $\sigma$-Algebras}
\label{sec:onto_sigma}

The central challenge in bias formalization is identifying the precise set of events that encode
sensitive attributes and their proxies. In a probability space $(\Omega,\mathcal{F},\mathbb{P})$, fairness
requires that a model’s output $Y$ be independent of a sub-$\sigma$-algebra $\mathcal{G} \subseteq \mathcal{F}$
that captures all “unsettling information.” The difficulty lies in systematically and completely
specifying $\mathcal{G}$: naive approaches that manually list features are ad hoc and prone to
omitting subtle proxies.

\subsection{Integrating Ontologies and $\sigma$-Algebras}

The central challenge in bias formalization is identifying the precise set of 
events that encode sensitive attributes and their proxies. In our setting, 
the sample space $\Omega$ is fixed as the finite set of data subjects 
(e.g., loan applicants or database rows). We equip $\Omega$ with the 
discrete $\sigma$-algebra $F = 2^{\Omega}$ and a probability measure $P$ 
(e.g., the empirical distribution over rows). For each sensitive or proxy 
concept $C$, entailment in the ontology allows us to identify the subset of 
$\Omega$ that falls under $C$. This guarantees that all such events are 
measurable, since $F$ contains all subsets of $\Omega$. 

\begin{definition}[Ontology-Guided $\sigma$-Algebra of Bias]
Let $O=(C,R,I,A)$ be an OWL~2-QL ontology with TBox $T$ and ABox $A$.
Let $S_{\text{onto}}\subseteq C$ contain every concept name explicitly
declared as a sensitive or proxy attribute in $O$.

Fix a countable sample space $\Omega$ (e.g., the finite set of
data-subject identifiers or database rows). Equip $\Omega$ with the
discrete $\sigma$-algebra $F=2^{\Omega}$ and a probability measure $P$.
For each concept $C\in S_{\text{onto}}$ define its extension
\[
  \llbracket C\rrbracket = \{\omega\in\Omega \mid
  O \models C(\omega)\},
\]
where the entailment relation $\models$ is evaluated in OWL~2-QL
(i.e., by rewriting into a SQL query over the ABox).

The \emph{ontology-guided $\sigma$-algebra of bias} is
\[
  \mathcal{G}_O =
    \sigma\!\bigl(\{\llbracket C\rrbracket \mid C\in S_{\text{onto}}\}\bigr)
    \subseteq F.
\]
Because $\Omega$ is finite, $\mathcal{G}_O$ is finite, uniquely
determined by the ABox, and computable in AC$_0$ data complexity.
\end{definition}

Intuitively, $\mathcal{G}_O$ is the smallest $\sigma$-algebra on $\Omega$ 
that makes the class extensions of all sensitive and proxy concepts measurable. 
Because $\Omega$ is finite, the construction is unique and computable by 
standard OWL~2-QL query answering. This shifts the fairness specification 
problem from ad hoc feature selection to a principled and auditable pipeline: 
sensitive attributes are declared in the ontology, compiled into measurable 
events via entailment, and collected into the bias $\sigma$-algebra.

\paragraph{Why Ontology-Guided $\sigma$-Algebras are Robust.}
This integration offers three decisive advantages:

\begin{enumerate}
  \item \textbf{Completeness.} Ontological axioms in the TBox allow 
  systematic discovery of proxy attributes through logical inference. 
  For example, an axiom such as
  \[
    \textsf{livesInZIP}(x,y) \wedge \textsf{Redlined}(y) 
    \;\Rightarrow\; \textsf{ProxyForRace}(x)
  \]
  ensures that $\mathcal{G}_O$ includes not only explicit sensitive 
  attributes but also indirect dependencies like ZIP~$\to$~race, 
  even when such correlations are not obvious in the raw data. 

  \item \textbf{Transparency and Auditability.} The TBox serves as a 
  machine-executable fairness policy. Regulators or auditors can inspect 
  axioms to verify which attributes and proxies generate measurable events 
  $\llbracket C\rrbracket$, making the fairness criterion both human-readable 
  and certifiable.

  \item \textbf{Mathematical Rigor.} By defining $\mathcal{G}_O$ as a 
  true $\sigma$-algebra generated by ontology-specified events 
  $\llbracket C\rrbracket$, subsequent fairness operations 
  (e.g., constructing $Y \perp\!\!\!\perp \mathcal{G}_O$) are applied to a 
  formally closed and comprehensive collection of events. This removes 
  arbitrariness in feature selection and guarantees provable independence.
\end{enumerate}

The ontology-guided $\sigma$-algebra $\mathcal{G}_O$ therefore transforms the 
informal task of “finding sensitive proxies” into a principled construction 
grounded in logic and measure theory. It provides a complete, transparent, 
and mathematically robust basis for certifiable fairness, bridging symbolic 
knowledge representation and probabilistic independence.

\subsection{Why Ontology-Guided $\sigma$-Algebras are Robust}
This integration offers three decisive advantages:

\begin{enumerate}
    \item \textbf{Completeness:} Ontological axioms in $\mathcal{T}$ allow systematic discovery of
    proxy attributes through logical inference. For example, axioms of the form
    \[
        \texttt{livesInZIP}(x,y) \wedge \texttt{Redlined}(y) \Rightarrow \texttt{ProxyForRace}(x)
    \]
    ensure that $\mathcal{G}_{\mathcal{O}}$ captures indirect dependencies (ZIP code $\to$ race)
    that may not be evident through correlation analysis alone.

    \item \textbf{Transparency and Auditability:} The TBox serves as a machine-executable
    fairness policy. Regulators or auditors can directly inspect axioms to verify which
    attributes and proxies generate $\mathcal{G}_{\mathcal{O}}$, making the fairness criterion
    both human-readable and certifiable.

    \item \textbf{Mathematical Rigor:} By defining $\mathcal{G}_{\mathcal{O}}$ as a true
    $\sigma$-algebra generated by ontology-specified events, subsequent fairness operations
    (e.g., constructing $Y \perp\!\!\!\perp \mathcal{G}_{\mathcal{O}}$) are applied to a formally
    closed and comprehensive collection of events. This eliminates the arbitrariness of
    feature selection and guarantees provable independence.
\end{enumerate}

The ontology-guided $\sigma$-algebra $\mathcal{G}_{\mathcal{O}}$ transforms the informal task of
“finding sensitive proxies” into a principled construction grounded in logic and measure theory.
It provides a complete, transparent, and mathematically robust basis for certifiable fairness,
bridging symbolic knowledge representation and probabilistic independence.

\subsection{A Case Study: Ontology-Guided $\sigma$-Algebra in Loan Approval}

To demonstrate the robustness of our framework, we revisit the loan approval
problem, integrating the measure-theoretic and ontological perspectives.
Traditional debiasing methods often fail because they treat bias superficially—
removing a small set of correlated features while missing subtle proxies.
In contrast, our ontology-guided $\sigma$-algebra $\mathcal{G}_O$ formalizes
and exhaustively captures all sensitive and proxy information, ensuring rigorous
independence.

\paragraph{Scenario: The Proxy Problem.}
Suppose a bank’s loan approval system is trained without explicit sensitive
variables such as gender or race, but uses features including an applicant’s
residential ZIP code and prior educational institution. While neither feature is
directly sensitive, both are well-known proxies:

\begin{itemize}
  \item Certain ZIP codes in the United States are historically linked to racial
  segregation or socio-economic disadvantage.
  \item Educational institutions, such as historically Black colleges or
  single-gender institutions, carry implicit demographic information about
  applicants.
\end{itemize}

Naive debiasing methods that simply drop the ZIP code feature are insufficient:
the model may still exploit correlations through the educational institution or
other variables. The bias ``flows’’ through alternative proxies.

\paragraph{Ontology-Guided Specification.}
We construct an ontology $O = (C,R,I,A)$ to formally encode this domain
knowledge:

\begin{itemize}
  \item Classes ($C$): \textsf{LoanApplicant}, \textsf{UrbanArea},
  \textsf{SuburbanArea}, \textsf{HistoricallyBlackCollege},
  \textsf{StateUniversity}, \textsf{ProxyForRace}, \textsf{ProxyForSES}.
  \item Roles ($R$): \textsf{livesIn} (applicant $\to$ area),
  \textsf{attended} (applicant $\to$ institution),
  \textsf{hasCreditScore} (applicant $\to \mathbb{R}$).
  \item Individuals ($I$): specific applicants such as \textsf{JohnDoe},
  \textsf{JaneSmith}.
  \item Axioms ($A$): logical rules linking classes and roles to sensitive
  attributes. For example:
  \[
    \textsf{attended}(x,y) \wedge y \in \textsf{HistoricallyBlackCollege}
    \;\Rightarrow\; x \in \textsf{ProxyForRace},
  \]
  \[
    \textsf{livesIn}(x,z) \wedge z \in \textsf{UrbanArea}
    \wedge \textsf{MedianIncome}(z) < 30{,}000
    \;\Rightarrow\; x \in \textsf{ProxyForSES}.
  \]
\end{itemize}

\paragraph{Ontology-Guided $\sigma$-Algebra.}
For each sensitive or proxy concept $C \in S_{\text{onto}}$, we compute its
extension
\[
  \llbracket C \rrbracket = \{\omega \in \Omega \mid O \models C(\omega)\}.
\]
The ontology-guided $\sigma$-algebra of bias is then
\[
  \mathcal{G}_O = \sigma\!\bigl(\{\llbracket C \rrbracket \mid C \in S_{\text{onto}}\}\bigr).
\]

In this example, $\mathcal{G}_O$ includes events such as:
\begin{itemize}
  \item the set of applicants living in ZIP codes designated as socio-economic
  proxies,
  \item the set of applicants who attended institutions correlated with race or
  gender,
  \item all Boolean combinations of such events.
\end{itemize}

\paragraph{Robustness of the Approach.}
This ontology-guided construction ensures:

\begin{enumerate}
  \item \textbf{Completeness:} All sensitive and proxy variables entailed by
  the ontology are automatically included in $\mathcal{G}_O$. This guards
  against residual bias hidden in indirect correlations.

  \item \textbf{Transparency:} The TBox axioms explicitly document the fairness
  policy, making $\mathcal{G}_O$ human-readable and auditable.

  \item \textbf{Provable Fairness:} By constructing a fair representation
  $Y \in L^2(\Omega,F,P)$ such that
  \[
    Y \;\perp\!\!\!\perp\; \mathcal{G}_O,
    \quad
    \mathbb{E}\!\left[\|X-Y\|^2\right]
    = \inf_{Z \perp\!\!\!\perp \mathcal{G}_O}
      \mathbb{E}\!\left[\|X-Z\|^2\right],
  \]
  We obtain certifiable independence from all sensitive and proxy information.
\end{enumerate}

\begin{figure}[H]
\centering
\begin{tikzpicture}[
    node distance=1.5cm,
    every node/.style={rectangle, rounded corners, draw, fill=gray!10, align=center, minimum width=5cm, minimum height=1.0cm},
    arrow/.style={-Stealth, thick}
]
\node[fill=pink!20] (A) {Ontology \\ $O = (C, R, I, A)$};
\node[below=of A] (B) {Sensitive/Proxy Classes \\ $S_{\text{onto}} \subseteq C$};
\node[below=of B] (C) {Events \\ $\llbracket C \rrbracket = \{\omega \in \Omega \mid O \models C(\omega)\}$};
\node[below=of C, fill=blue!20] (D) {Bias $\sigma$-Algebra \\ $\mathcal{G}_O = \sigma(\{\llbracket C \rrbracket\})$};
\node[below=of D] (E) {Probability Space \\ $(\Omega, F, P),\;\; \mathcal{G}_O \subseteq F$};
\node[below=of E] (F) {Fair Representation \\ $Y \perp\!\!\!\perp \mathcal{G}_O$};
\node[below=of F] (G) {Optimal $Y$ \\ $\arg\min_{Z \perp\!\!\!\perp \mathcal{G}_O} \mathbb{E}[\|X - Z\|^2]$};
\node[below=of G] (H) {Debiased Decisions \\ (e.g., Loan Approval)};

\node[right=2cm of D, fill=gray!20, dashed, inner sep=8pt, align=left, minimum width=6cm] (R) {
    \textbf{Robustness Advantages:} \\
    1. Completeness: Infers hidden proxies (e.g., ZIP $\to$ race). \\
    2. Transparency: Axioms form an auditable fairness policy. \\
    3. Mathematical rigor: $\sigma$-algebra closed under $\cup, \cap, \complement$.
};

\draw[arrow] (A) -- node[right, font=\footnotesize, align=left, draw=none, fill=none] {TBox + ABox assertions} (B);
\draw[arrow] (B) -- node[right, font=\footnotesize, draw=none, fill=none] {Logical inference $O \models \alpha$} (C);
\draw[arrow] (C) -- node[right, font=\footnotesize, draw=none, fill=none] {Generate $\sigma$-algebra} (D);
\draw[arrow] (D) -- (E);
\draw[arrow] (E) -- node[right, font=\footnotesize, draw=none, fill=none] {Enforce independence} (F);
\draw[arrow] (F) -- node[right, font=\footnotesize, draw=none, fill=none] {$L^2$ minimization} (G);
\draw[arrow] (G) -- node[right, font=\footnotesize, draw=none, fill=none] {Application} (H);
\draw[arrow] (D.east) -- (R.west);
\end{tikzpicture}
\caption{Ontology-guided $\sigma$-algebra construction and robustness in AI
fairness. Sensitive and proxy classes $S_{\text{onto}}$ are identified in the
ontology, extended to measurable events $\llbracket C \rrbracket$, and generate
the bias $\sigma$-algebra $\mathcal{G}_O \subseteq F$. A fair representation $Y$
is then constructed such that $Y \perp\!\!\!\perp \mathcal{G}_O$ and is optimal
in $L^2$, ensuring debiased loan approval decisions while guaranteeing
completeness, transparency, and mathematical rigor.}
\label{fig:ontology-sigma-algebra}
\end{figure}
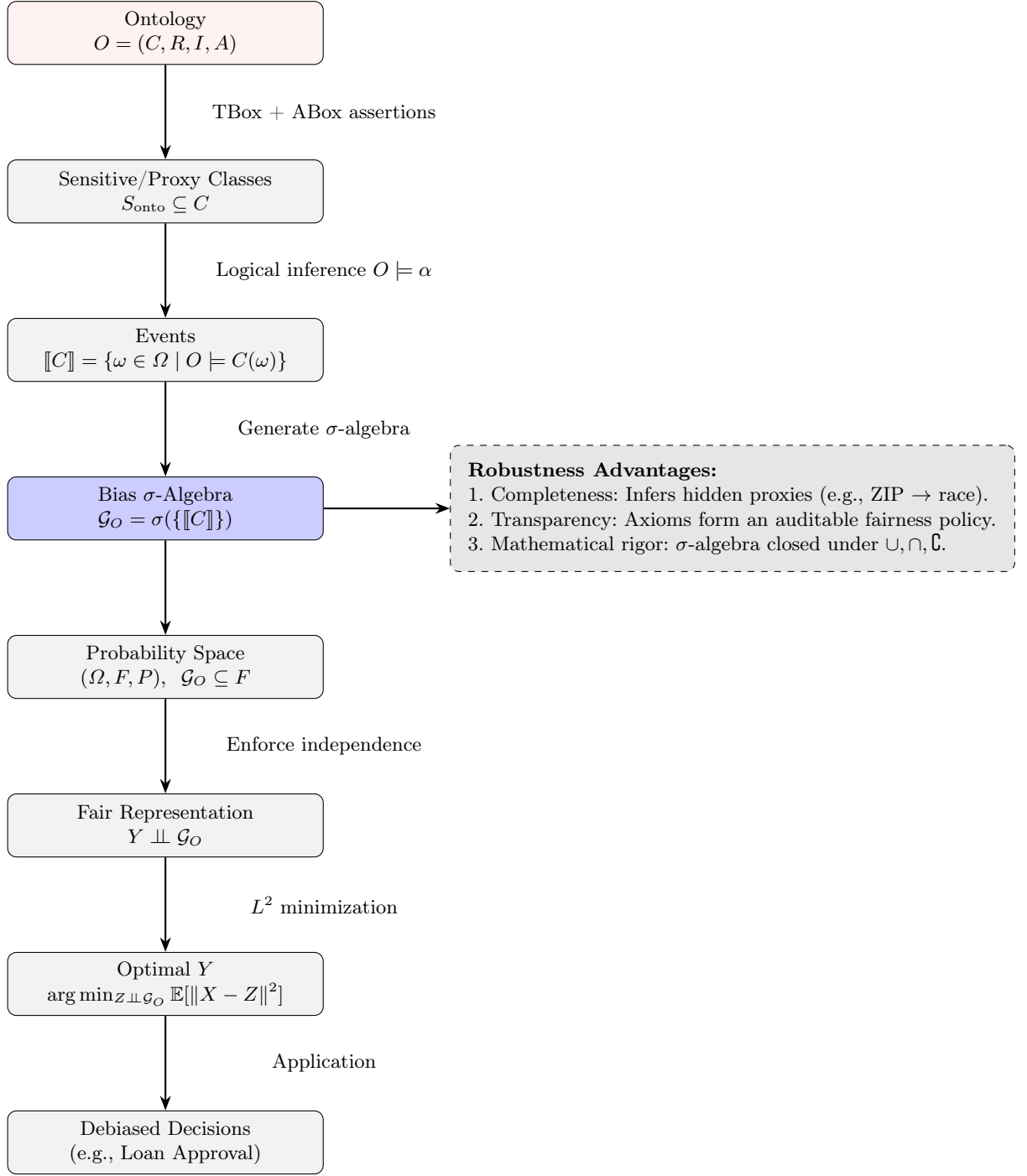

The ontology-guided $\sigma$-algebra $\mathcal{G}_O$ thus provides a rigorous,
auditable, and mathematically grounded way to capture the complete structure
of bias in loan approval, ensuring that fairness is enforced not only for
explicitly sensitive variables but also for their hidden proxies.

\section{Implementation Blueprint}
\label{sec:imp}

This section outlines a prospective engineering plan that translates the
theoretical framework of Sections \S\ref{sec:sigma}, \S\ref{sec:onto} and
\S\ref{sec:onto_sigma} into a reproducible pipeline. The objective is to enable
practitioners to move from \emph{ethical intent} (encoded in an OWL~2-QL
ontology) to \emph{provably fair decisions} with minimal friction. At this
stage, the blueprint remains conceptual and intended as a roadmap for future
implementation, leveraging established tools like the open-source OWL reasoner
HermiT~\cite{glimm2014hermit} and benchmark fairness datasets like
COMPAS~\cite{fabris2022algorithmic} to ensure practical grounding.

\subsection{End-to-End Pipeline}
\label{sub:pipeline}

\begin{figure}[H]
  \centering
  \begin{tikzpicture}[
    box/.style={
      rectangle,
      draw,
      rounded corners,
      minimum height=3em,
      minimum width=3.0cm,
      text width=3.0cm,
      align=center,
      fill=blue!10,
      font=\small
    },
    arrow/.style={-Stealth, thick}
  ]
    \node[box] (ontology) at (0,0) {Ontology Authoring\\(OWL~2-QL axioms, Prot\'{e}g\'{e}, HermiT)};
    \node[box, right=0.75cm of ontology] (sigma) {$\sigma$-Algebra Generator\\(Mask matrix $M$, entailment queries $O \models C(\omega)$)};
    \node[box, right=0.75cm of sigma] (fair) {Fair Representation Engine\\(Optimal transport, PyTorch)};
    \node[box, right=0.75cm of fair] (audit) {Certification \& Audit\\(JSON-LD certificate, HSIC test)};

    \draw[arrow] (ontology) -- (sigma);
    \draw[arrow] (sigma) -- (fair);
    \draw[arrow] (fair) -- (audit);
  \end{tikzpicture}
  \caption{Planned implementation pipeline: sensitive and proxy classes
  $S_{\text{onto}}$ are declared in the ontology, compiled into measurable
  events $\llbracket C \rrbracket$, and aggregated into the bias
  $\sigma$-algebra $\mathcal{G}_O$. Optimal transport then constructs a fair
  representation $Y \perp\!\!\!\perp \mathcal{G}_O$, and certification provides
  auditability.}
  \label{fig:pipeline}
\end{figure}
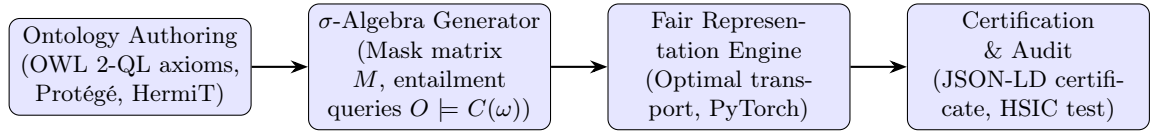

Figure~\ref{fig:pipeline} illustrates the four macro-stages that form our
reference implementation design:
\begin{enumerate}
  \item \textbf{Ontology Authoring} (\S\ref{sub:authoring})
  \item \textbf{$\sigma$-Algebra Generator} (\S\ref{sub:sigma-gen})
  \item \textbf{Fair Representation Engine} (\S\ref{sub:ot-engine})
  \item \textbf{Certification \& Audit} (\S\ref{sub:audit})
\end{enumerate}

\subsection{Ontology Authoring Guidelines}
\label{sub:authoring}

We propose structuring the ontology in three layers:
\begin{itemize}
  \item \textbf{Upper Layer:} Generic fairness vocabulary 
  (e.g., \texttt{SensitiveAttribute}, \texttt{ProxyAttribute}).
  \item \textbf{Domain Layer:} Mid-level domain concepts such as 
  \texttt{LoanApplicant}, \texttt{ZIPCode}, \texttt{HBCU}.
  \item \textbf{Policy Layer:} Project-specific axioms, e.g.,
  \[
    \texttt{livesIn}(x,\texttt{zip}) \wedge (\texttt{MedianIncome}(\texttt{zip}) < 30{,}000)
    \;\longrightarrow\; \texttt{ProxyForSES}(x).
  \]
\end{itemize}
The final ontology is serialized in RDF/XML and passed as input to the
$\sigma$-Algebra Generator.

\subsection{$\sigma$-Algebra Generator}
\label{sub:sigma-gen}

\paragraph{\textbf{Inputs:}}
(1) OWL~2-QL ontology file, (2) dataset $\mathcal{D}=\{\omega_i\}_{i=1}^N$, (3)
mapping from ontology individuals to dataset rows.

\paragraph{\textbf{Output:}}
A binary mask matrix $M \in \{0,1\}^{N \times k}$ where
\[
  M_{i,j} = 1 \iff O \models C_j(\omega_i), \quad C_j \in S_{\text{onto}}.
\]
Each column corresponds to an event $\llbracket C_j \rrbracket$, and the set
$\{\llbracket C_j \rrbracket \mid C_j \in S_{\text{onto}}\}$ generates the bias
$\sigma$-algebra $\mathcal{G}_O$.

\paragraph{Complexity:}
Computing $M$ reduces to $|S_{\text{onto}}|$ SPARQL \texttt{ASK} queries per row,
i.e., $\mathcal{O}(N \cdot |S_{\text{onto}}|)$. With columnar storage
(Parquet/ORC) and vectorized query engines, datasets with up to $10^7$ rows
should be processable on a single multi-core node within practical time limits.

\subsection{Fair Representation Engine}
\label{sub:ot-engine}

We plan to implement the Delbaen--Majumdar optimal-transport solver in PyTorch.
Given $M$ and the original feature matrix $X \in \mathbb{R}^{N \times d}$, the
procedure will:
\begin{enumerate}
  \item Estimate conditional expectations $\hat{\mu}_g = \mathbb{E}[X \mid g]$
  for each $g$ in the partition induced by $M$ (intersections of
  $\llbracket C_j \rrbracket$).
  \item Solve the discrete optimal-transport problem to obtain coupling $\pi^\star$.
  \item Map $X$ to $Y$ such that $Y \perp\!\!\!\perp \mathcal{G}_O$.
\end{enumerate}

\begin{algorithm}[h]
\small
\caption{Planned Optimal-Transport Fair Projection}
\label{alg:ot}
\begin{algorithmic}[1]
\Require $X \in \mathbb{R}^{N \times d}$, mask matrix $M$, regularization $\varepsilon > 0$
\State Build partition $\{\Omega_g\}_{g \in \mathcal{G}_O}$ from unique rows of $M$
\State Compute $\mu_g = \tfrac{1}{|\Omega_g|}\sum_{i \in \Omega_g} X_i$ for all $g$
\State Solve entropic OT:
$\pi^\star = \arg\min_{\pi} \sum_{i,j} \pi_{ij} \|X_i - \mu_{g_j}\|^2 + \varepsilon H(\pi)$
\State Return $Y_i = \sum_j \pi^\star_{ij}\,\mu_{g_j}$
\end{algorithmic}
\end{algorithm}

\paragraph{Complexity:}
Naively $\mathcal{O}(N^2 d)$, but Sinkhorn iterations on GPU are expected to
reduce wall-clock time to seconds for $N \approx 10^6$.

\subsection{Certification \& Audit}
\label{sub:audit}

\paragraph{Planned Certificate Contents:}
Each model would be accompanied by a JSON-LD certificate containing:
\begin{itemize}
  \item SHA-256 hash of the ontology file,
  \item hash of the mask matrix $M$,
  \item reconstruction error $\|X-Y\|^2_F / N$,
  \item independence test $p$-value via HSIC~\cite{wang2021learning}.
\end{itemize}

\paragraph{Regulatory Audit Hook:}
Third-party auditors could re-run the pipeline with the disclosed ontology and
dataset (e.g., COMPAS~\cite{fabris2022algorithmic}) to verify that the generated
$\sigma$-algebra $\mathcal{G}_O$ matches the training-time one.

\subsection{Deployment Footprint}

We envisage packaging the system as:
\begin{itemize}
  \item A \texttt{docker-compose} stack (OWL reasoner like HermiT, Spark job
  for $M$, PyTorch service for OT).
  \item Helm charts for Kubernetes with autoscaling GPU nodes.
  \item A minimal CPU-only fallback for development environments.
\end{itemize}

This modular design balances scalability with accessibility. Performance
estimates (e.g., end-to-end latency for multi-million-row datasets) remain to be
validated in a full implementation.

\section{Conclusion}
\label{sec:conclusion}

This work presents a novel framework that bridges the gap between ethical
declarations and mathematical guarantees in AI fairness by integrating ontology
engineering with measure-theoretic optimal transport. Our approach addresses a
fundamental limitation in existing bias mitigation techniques: the inability to
systematically identify and eliminate all forms of sensitive information,
including subtle proxies that enable discriminatory decision-making through
seemingly neutral features.

The theoretical foundation of our framework rests on three key innovations.
First, we formalize algorithmic bias as a structural property of a model’s
learned $\sigma$-algebra, providing a mathematically rigorous characterization
that extends beyond simple correlation-based measures. Second, we demonstrate
how OWL~2-QL ontologies can be systematically compiled into $\sigma$-algebras by
defining class extensions $\llbracket C \rrbracket = \{\omega \in \Omega \mid O
\models C(\omega)\}$ for sensitive and proxy concepts. This construction
ensures that $\mathcal{G}_O = \sigma(\{\llbracket C \rrbracket\})$ captures both
explicitly declared sensitive attributes and their proxies derived by logical
inference. Third, we establish that Delbaen--Majumdar optimal transport provides
the unique solution for constructing fair representations that are provably
independent of biased information while minimizing information loss in the
$L^2$-norm.

Our loan approval case study illustrates the practical significance of this
approach. Traditional debiasing methods that merely remove explicitly sensitive
features fail to address the ``proxy problem,'' where correlated variables such
as ZIP codes or educational institutions continue to encode protected
characteristics. By contrast, our ontology-guided $\sigma$-algebra
$\mathcal{G}_O$ systematically identifies these indirect pathways to bias,
ensuring that the resulting fair representation $Y$ satisfies
$Y \perp\!\!\!\perp \mathcal{G}_O$---a guarantee of exact statistical
independence rather than mere decorrelation.

The framework’s robustness stems from three critical advantages that distinguish
it from existing approaches. \emph{Completeness} is achieved through ontological
reasoning that systematically discovers proxy relationships, preventing
residual bias from hidden correlations. \emph{Transparency} emerges from the
explicit axiomatization of fairness policies in machine-readable form, enabling
regulatory audit and public scrutiny. \emph{Mathematical rigor} is ensured
through the closure properties of $\sigma$-algebras and optimal transport,
providing certifiable guarantees rather than heuristic approximations.

While we have established the theoretical foundations and provided a detailed
implementation blueprint, several important directions warrant future
investigation. Empirical validation across diverse domains and datasets remains
essential to demonstrate the framework’s practical effectiveness and
computational scalability. The proposed pipeline, leveraging OWL~2-QL reasoning
and PyTorch-based optimal transport solvers, requires full implementation and
performance evaluation on large-scale datasets. Additionally, the framework’s
behavior under distribution shift and its integration with existing MLOps
pipelines merit careful study.

The broader implications of this work extend beyond technical considerations to
fundamental questions about the role of AI in society. As AI systems
increasingly influence consequential decisions in lending, hiring, healthcare,
and criminal justice, the need for mathematically grounded fairness guarantees
becomes paramount. Our framework provides a pathway from informal ethical
commitments to formal mathematical constraints, offering a foundation for
building AI systems that are not merely performant but demonstrably equitable.

By ensuring that fairness constraints are both comprehensive and mathematically
verifiable, the integration of symbolic reasoning through ontologies with
probabilistic independence through optimal transport contributes to the broader
goal of developing trustworthy AI systems that serve all members of society
equitably.

%
%
%
\bibliographystyle{splncs04}
\bibliography{ref}
\end{document}